\documentclass[letterpaper]{article}
\usepackage{aaai20}
\usepackage{times}
\usepackage{helvet}
\usepackage{courier}
\usepackage[hyphens]{url}
\usepackage{graphicx}
\usepackage{graphicx}
\usepackage{amsmath}
\usepackage{amssymb}
\usepackage{todonotes}
\usepackage{booktabs}
\usepackage{multirow}
\usepackage[normalem]{ulem}

\newcommand{\R}{\mathbb{R}}

\title{An Evaluation of Knowledge Graph Embeddings for Autonomous Driving Data: Experience and Practice}

\author{Ruwan Wickramarachchi\textsuperscript{1},
Cory Henson\textsuperscript{2}, and 
Amit Sheth\textsuperscript{1}\\
\textsuperscript{1}{Artificial Intelligence Institute, University of South Carolina, Columbia, SC, USA}\\
ruwan@email.sc.edu, amit@sc.edu\\
\textsuperscript{2}{Bosch Research and Technology Center, Pittsburgh, PA, USA}\\
cory.henson@us.bosch.com 
}

\begin{document}
\maketitle
\begin{abstract}
%

The autonomous driving (AD) industry is exploring the use of knowledge graphs (KGs) to manage the vast amount of heterogeneous data generated from vehicular sensors. The various types of equipped sensors include video, LIDAR and RADAR.  Scene understanding is an important topic in AD which requires consideration of various aspects of a scene, such as detected objects, events, time and location. Recent work on knowledge graph embeddings (KGEs) - an approach that facilitates neuro-symbolic fusion - has  shown to improve the predictive performance of machine learning models. With the expectation that neuro-symbolic fusion through KGEs will improve scene understanding, this research explores the generation and evaluation of KGEs for autonomous driving data. We also present an investigation of the relationship between the level of informational detail in a KG and the quality of its derivative embeddings. By systematically evaluating KGEs along four dimensions -- i.e. quality metrics, KG informational detail, algorithms, and datasets -- we show that (1) higher levels of informational detail in KGs lead to higher quality embeddings, (2) type and relation semantics are better captured by the semantic transitional distance-based TransE algorithm, and (3) some metrics, such as coherence measure, may not be suitable for intrinsically evaluating KGEs in this domain. Additionally,  we also present an (early) investigation of the usefulness of KGEs for two use-cases in the AD domain.

\end{abstract}

\section{Introduction} 
\label{sec:intro}

Forecasters predict that fully autonomous vehicles could be commercially available in the next few years; and within the next few decades (i.e. 2040) half of all vehicles sold and 40 percent of vehicle travel could be autonomous \cite{litman2019}. While racing to realize this vision, the automotive industry is investing heavily into machine learning and other relevant AI technologies. To meet the increasing data demands of ML algorithms, fleets of vehicles are now deployed in multiple cities around the world and collecting massive amounts of data. These vehicles are equipped with various types of heterogeneous sensors, including – but not limited to – video, LIDAR, and RADAR.

To manage these vast amounts of automotive sensor data, companies are beginning to experiment with the use of KGs. In other industries, and for many years, KGs have proven invaluable for helping to manage data stored within enterprise data lakes, which are growing rapidly in popularity. More specifically, KGs help to enable the principles of FAIR data – i.e. findability, accessibility, interoperability, and re-use – across an enterprise.

Current research into the topic of neuro-symbolic fusion\footnote{https://www.digitaltrends.com/cool-tech/neuro-symbolic-ai-the-future/}  \cite{nickel2015review,shadesof}, however, is beginning to point to new and exciting uses of KGs. Essentially, KGs are a key source of high-quality domain knowledge and KGE technology is now enabling ML algorithms to more directly access this knowledge. Recent studies have already shown that the use of KGEs leads to improved performance and predictive capabilities \cite{chen2017multilingual,wang2019multi,myklebust2019knowledge}. For this reason, we believe that neuro-symbolic fusion through KGEs may provide valuable knowledge needed to improve scene understanding for autonomous driving.

In this paper, we share our experience with generating and evaluating KGEs for the AD domain. To the best of our knowledge, this is the first attempt of its kind in this domain. Our investigation begins with two popular benchmark datasets from Aptiv and Lyft. From each of these datasets we generate multiple KGs with varying degrees of informational detail. The generated KGs focus on representing the various scenes, or situations, that an autonomous vehicle encounters on the road. The purpose of creating KGs with varying degrees of detail is to enable an examination of the relationship between KG detail and the quality of derivative embeddings. Each KG is then translated into a set of KGEs, each derived from one-of three popular embedding algorithms; including TransE \cite{bordes2013translating}, RESCAL \cite{nickel2011three}, and HoLE \cite{nickel2016holographic}. Finally, the quality of each KGE is systematically evaluated based on the framework proposed in \cite{alashargi2019metrics}. 

Our analysis of evaluating the KGEs along four dimensions -- i.e. quality metrics, KG informational detail, algorithms, and datasets -- leads to some interesting findings. First, we show that KGE quality significantly improves as the informational detail of a KG increases. Second, focusing on the evaluation measures, we report that some of the metrics such as the coherence measure may not be suitable to evaluate KGEs in this domain. When considering the effectiveness of KGE algorithms, we identify that the semantic transitional distance-based TransE algorithm captures type and relational semantics better than algorithms from the class of semantic matching-based models. It is interesting to note that these findings are consistent across the evaluations on two datasets. Finally, we report preliminary observations on using KGEs for two use cases from the AD domain. Specifically, we demonstrate how the scene/sub-scene understanding was improved as KG informational detail was increased, and how KGEs can be used to compute scene similarity.

The two primary contributions of this paper include: (1) a demonstration of the process of creating and evaluating KGEs for AD data, and (2) an (early) examination of the relationship between KG detail and the quality of KGEs. 
In Section 2, we discuss the construction of KGs from the benchmark automotive driving datasets. The translation of KGs to KGEs is explained in Section 3, while Section 4 focuses on their evaluation. Details of the technology used,  as well as related work, will be discussed in each individual section. An investigation on the usefulness of semantics in the AD domain is discussed in Section 5. Finally, in Section 6 we conclude with a summary of our overall results and directions for future research.

\section{Scene Knowledge Graphs}
\label{sec:scene-kgs}
To evaluate the KGEs for the AD domain, several KGs were created based on two popular benchmark datasets; NuScenes from Aptiv  \cite{Caesar2019nuScenesAM} and Lyft-Level5 from Lyft \cite{lyft2019}. To annotate the data from the datasets, a scene ontology was used.

\subsection{Composition of the Datasets}
Both the NuScenes and Lyft datasets follow a similar structure. NuScenes, for example, is divided into a set of 20 second driving segments/scenes, with $\sim$40 samples/sub-scenes per segment (i.e. one sample/sub-scene every 0.5 seconds). Each 20 second segment is associated with a temporal interval and spatial area, while each sample is associated with the data collected at a specific temporal instant (i.e. timestamp) and spatial coordinates. 

The NuScenes dataset contains 850 driving segments with 34,149 samples. Each object and event detected in a sample is associated with one of 23 categories\footnote{https://www.nuscenes.org/data-annotation}.

The Lyft dataset contains 180 driving segments with 22,680 samples. Lyft has only 9 object and event categories that are used for annotating samples.

\subsection{Scene Ontology}
A scene is described as an observable volume of time and space \cite{henson2019iswc}. In the AD domain, a scene depicts a situation encountered by a vehicle. A few examples may include a vehicle stopped at a traffic light, cruising on the highway, or crashing into another vehicle. The concept of scene acts as the polestar with which all information about the vehicle, and its situation, are integrated. More specifically, a scene may include information about time and location, the occurring events, and the participating objects.

A scene may also include sub-scenes. For example, consider a vehicle driving for 20 seconds on a highway. This drive may be represented as a single scene. However, during this drive the vehicle may encounter several different situations, each of which may also be represented as a scene. 

\begin{figure}[h]
\centerline{\includegraphics[width=0.3\textwidth]{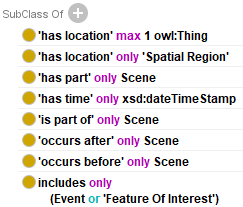}}
\caption {Formal definition of a scene, as defined by the Scene Ontology}
\label{fig:scene-defn}
\end{figure}

Formally, Figure \ref{fig:scene-defn} shows the properties associated with a scene (depicted in Protege\footnote{https://protege.stanford.edu/}). 

\begin{figure}[h]
\centerline{\includegraphics[width=0.25\textwidth]{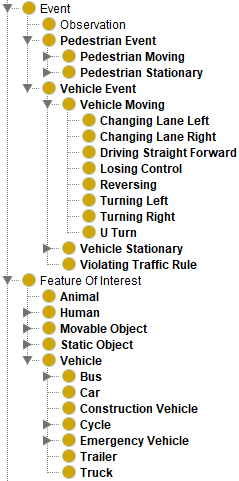}}
\caption {Subset of events and features-of-interest contained in the Scene Ontology.}
\label{fig:fois-and-events}
\end{figure}

A subset of the events and features-of-interest (i.e. objects) represented within the Scene Ontology is shown in Figure \ref{fig:fois-and-events}. For this work, the Scene Ontology has been extended to subsume all concepts found in both the NuScenes and Lyft datasets.

\subsection{Informational Detail of KGs }

For each dataset, three distinct KGs are generated with differing levels of informational detail. It should be noted that the level of informational detail for each KG refers to the inclusion of additional information about scenes. The three levels include: (1) a base KG, (2) a KG with inferred \textit{type} relations for objects/events, and (3) a KG with additional \textit{includes} relations between scenes and object/events. It should be noted that this additional information does not correspond to an increase in the logical expressivity of the KGs. It is also worth mentioning that each KG includes the Scene Ontology along with the facts derived from each dataset.

\begin{figure}[!htbp]
\centerline{\includegraphics[width=0.4\textwidth]{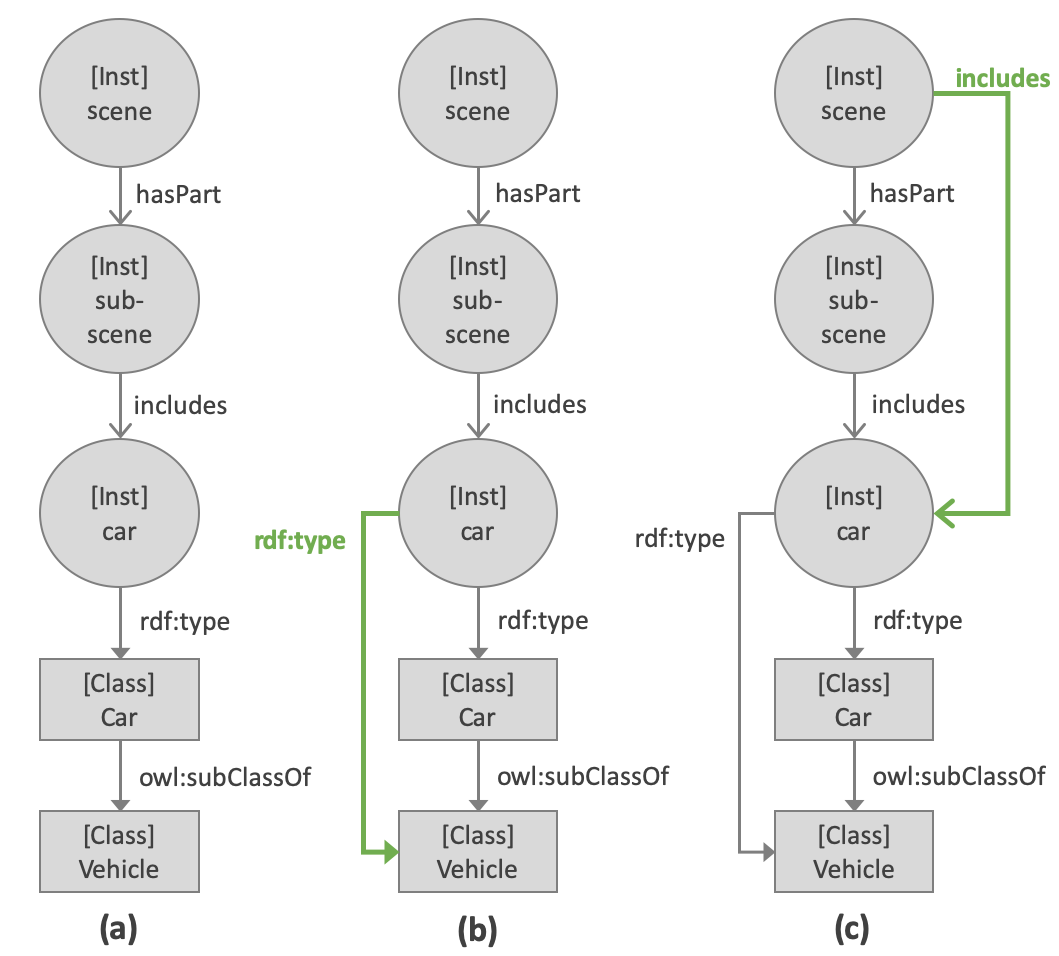}}
\caption{Example KGs: (a) Base KG, (b) KG w/ inferred types, and (c) KG w/ include path}
\label{fig:kg-patterns}
\end{figure}

\subsubsection{Base KG}
Within the Base KG, each 20 second segment is instantiated as a scene, and each sample is instantiated as a sub-scene. The scene representing a 20 second segment is associated with a temporal interval (of 20 seconds) and a spatial location (e.g. a city). This scene is also associated with a set of sub-scenes, representing the samples. Each sub-scene is associated with a temporal instant, spatial point, and the objects and events that participate in the scene. See Figure \ref{fig:kg-patterns}(a) for an example.

\subsubsection{KG with Inferred Types}
In the Base KG, objects and events are explicitly typed to the most specific class possible. For example, an object instance representing a car is typed to the \textit{Car} class. Because the \textit{Car} class is a sub-class of \textit{Vehicle}, then the instance is also a type of \textit{Vehicle}. However, this knowledge is only implicit in the KG; implied by the semantics of the \texttt{owl:subClassOf} relation. To make this knowledge explicit, a reasoner is used to infer all implied types for each object and event instance. See Figure \ref{fig:kg-patterns}(b) for an example.
\\
\\Example RDF (new triples proceeded by $\xrightarrow{}$)
\\:inst-scene rdf:type scene:Scene . 
\\:inst-scene scene:hasPart  :inst-sub-scene . 
\\:inst-sub-scene includes :inst-car .
\\:inst-car rdf:type scene:Car . 
\\$\xrightarrow{}$ \textit{:inst-car rdf:type scene:Vehicle .} 
\\$\xrightarrow{}$ \textit{:inst-car rdf:type scene:FeatureOfInterest .}

\subsubsection{KG with Include Paths}
Within the Base KG, objects and events are associated with sub-scenes derived from samples of a 20 second drive. The scene representing the entire drive is associated with these participating objects and events through a two-hop path.
\\
\\Example RDF 
\\:inst-scene scene:hasPart :inst-sub-scene . 
\\:inst-sub-scene scene:includes :inst-car . 
\\

In order to make a more direct association between a scene representing a drive and the detected objects and events, new \textit{includes} relations are added to the KG. See Figure \ref{fig:kg-patterns}(c)  for an example.
\\
\\Example RDF (new triples proceeded by $\xrightarrow{}$)
\\:inst-scene scene:hasPart :inst-sub-scene . 
\\:inst-sub-scene scene:includes :inst-car . 
\\$\xrightarrow{}$
\textit{:inst-scene scene:includes :inst-car .}
\\

Table \ref{tab:kg-stats} summarizes the statistics of the three KG versions we have generated.

\begin{table}[htbp]
\centering
\begin{tabular}{lcccc}
\cline{3-5}
 & \multicolumn{1}{c|}{} & \multicolumn{1}{c|}{\multirow{3}{*}{Base}} & \multicolumn{1}{c|}{\multirow{3}{*}{\begin{tabular}[c]{@{}c@{}}W/\\ inferred\\ types\end{tabular}}} & \multicolumn{1}{c|}{\multirow{3}{*}{\begin{tabular}[c]{@{}c@{}}W/\\ include\\ paths\end{tabular}}} \\
 & \multicolumn{1}{l|}{} & \multicolumn{1}{c|}{} & \multicolumn{1}{c|}{} & \multicolumn{1}{c|}{} \\
 & \multicolumn{1}{l|}{} & \multicolumn{1}{c|}{} & \multicolumn{1}{c|}{} & \multicolumn{1}{c|}{} \\ \hline
\multicolumn{1}{|c|}{\multirow{3}{*}{NuScenes}} & \multicolumn{1}{c|}{\# triples} & \multicolumn{1}{c|}{5.95M} & \multicolumn{1}{c|}{8.78M} & \multicolumn{1}{c|}{10.80M} \\ \cline{2-5} 
\multicolumn{1}{|c|}{} & \multicolumn{1}{c|}{\# entities} & \multicolumn{3}{c|}{2.11M} \\ \cline{2-5} 
\multicolumn{1}{|c|}{} & \multicolumn{1}{c|}{\# relations} & \multicolumn{3}{c|}{11} \\ \hline
 &  &  &  &  \\ \hline
\multicolumn{1}{|l|}{\multirow{3}{*}{Lyft}} & \multicolumn{1}{c|}{\# triples} & \multicolumn{1}{c|}{3.94M} & \multicolumn{1}{c|}{5.85M} & \multicolumn{1}{c|}{7.12M} \\ \cline{2-5} 
\multicolumn{1}{|l|}{} & \multicolumn{1}{c|}{\# entities} & \multicolumn{3}{c|}{1.33M} \\ \cline{2-5} 
\multicolumn{1}{|l|}{} & \multicolumn{1}{c|}{\# relations} & \multicolumn{3}{c|}{11} \\ \hline
\end{tabular}
\caption{Statistics of the three KG versions generated from the NuScenes and Lyft datasets}
\label{tab:kg-stats}
\end{table}

\section {KG Embeddings for AD Scenes}
The goal of learning embeddings from a KG is to represent the entities and relations in low-dimensional vector space while also maintaining the semantics contained in the KG. This transformation allows KGs to be more easily manipulated and used for downstream learning tasks (e.g. link prediction \cite{xiao2016one} and KG completion \cite{lin2015learning}). Vector representation of KGEs also allows background knowledge contained in KGs to be easily integrated with other input features of a machine learning model.

To select candidate algorithms for our experiments, we referred to the classification of KGE algorithms established by \cite{wang2017knowledge} and \cite{sharma2018towards}. KGE algorithms are categorized into two main classes: (1) Transitional distance-based algorithms and (2) Semantic Matching Models. For transitional distance-based algorithms (i.e. additive methods) the scoring function is composed of distance measures, and vector addition/subtraction is used to capture the vector interaction. For Semantic Matching Models (i.e. multiplicative methods) the scoring function is based on a similarity measure, and entity-relation-entity interaction is captured via a multiplicative score function. We initially selected one algorithm from each class. Specifically, we selected TransE from the former category and RESCAL from the latter category. RESCAL, however, has limitations in handling big KGs due to its high space and time complexity. As a result, we also included HolE into our experiments, which is a more space and memory-efficient successor of RESCAL.

\subsection{Preliminaries}
Next, the symbols and notations used throughout the paper are introduced, along with important details of the KGE algorithms.
\subsubsection{Notation:}
Given a set of entities $\mathcal{E}$ and set of relations $\mathcal{R}$, we define KG to be a set of triples ($h,r,t$), $\mathcal{T} = \mathcal{E} \times R \times (\mathcal{E} \cup \mathcal{L}$) where $\mathcal{L}$ is the set of literals. We consider $\mathcal{E} = \mathcal{C} \cup \mathcal{N}$ where $\mathcal{C}$ is set of concepts from the ontology and $\mathcal{N}$ is set of individuals. Lowercase bold characters represent vectors of an entity or relation and uppercase bold characters represent a set of vectors. For example, $\mathbf{e} \in \mathbf{E}$ is an embedding vector of $e \in \mathcal{E}$. Most of the embedding algorithms -- including some used in our evaluation -- generate vectors of dimension $d$ to represent entities $\mathbf{e} \in \R^{d}$ for $e \in \mathcal{E}$ and $\mathbf{r} = \R^{d}$ to represent relations $r \in \mathcal{R}$. However, some algorithms learn a projection matrix $M_r \in \R^{d \times d}$ to represent relations. The scoring function $\sigma :  \mathcal{E} \times \mathcal{R} \times \mathcal{E} \xrightarrow{} \R$ used in each algorithm is different. Transitional distance-based models use distance measures whereas semantic matching based methods use  similarity measures. The learning of embeddings involve optimizing parameter $\theta$ in the loss function $\mathcal{L} (\mathcal{T}, \mathcal{T'}, \theta$) where $\mathcal{T}$ is the set of positive triples and  $\mathcal{T'}$ is the set of corrupted triples. When considering time and space complexities of each algorithm, we consider $n =|E|$, $m = |R|$ and $n_t$ to be the number of training triples.

The details of each algorithm used are briefly discussed below;

\subsubsection{TransE}
TransE is considered the most representative of the translational distance-based class of algorithms. Given a triple ($h, r, t$), TransE represents $r$ as a translation vector from $h$ to $t$. Hence $\mathbf{h} + \mathbf{r} \approx \mathbf{t}$ when the triple ($h, r, t$) holds true. The scoring function of TransE $f_r$ is defined as the negative distance between $\mathbf{h} + \mathbf{r}$ and $\mathbf{t}$, and when the ($h, r, t$) holds, $f_r$ is expected to be large. 
\begin{equation}
f_r (h, t) =  - |\mathbf{h} + \mathbf{r} - \mathbf{t}|_{1/2}
\end{equation}

TransE is one of the most efficient KGE algorithms having $\mathcal{O} (n d + m d)$ space complexity and $\mathcal{O}(n_t d)$ time complexity. Despite it's benefits, TransE falls short in capturing 1-N, N-1 and N-N relations in KGs.

\subsubsection{RESCAL}
RESCAL belongs to the semantic matching/multiplicative class of KGE algorithms. RESCAL is an expressive model which takes into account the inherent structure in multi-relational KGs and captures complex patterns over multiple hops in the KG. It represents each relation $r$ as matrix $M_r$ that captures all the interaction between vectors $\mathbf{(h,t)}$ of the entities $h$ and $r$. The scoring function $f_r (h, t)$ is a bi-linear function which computes pairwise interaction of entities with respect to each relation $r$.  

\begin{equation}
    f_r = \mathbf{h}^T \mathbf{M}^r \mathbf{t} = \sum_{i=0}^{d-1}\sum_{j=0}^{d-1} [\mathbf{M}_r]_{ij} . [\mathbf{h}]_i . [\mathbf{t}]_j
\end{equation}

The main limitation of using RESCAL with big KGs is due to its high space and time complexity. RESCAL has $\mathcal{O} (n d + m d^2)$ space complexity and $\mathcal{O}(n_t d^2)$ time complexity.

\subsubsection{HolE}
HolE is an efficient successor of RESCAL and addresses the high space complexity of RESCAL while retaining the expressive power. HolE represents both entities and relations as vectors in $\R^d$. Given a triple ($h, r, t$), HolE first creates a compositional vector using circular correlation operation which aims at compressing the pairwise interaction;

\begin{equation}
    [\mathbf{h}\star \mathbf{t}] = \sum_{k=0}^{d-1} [\mathbf{h}]_k . [\mathbf{t}]_{(k+i)\mod d}
\end{equation}

The total score for a given fact is then calculated by using the function $f_r (h,t)$ that considers both the compositional vector and the relation vector. 

\begin{equation}
    f_r (h,t) = \mathbf{r}^T (\mathbf{h} \star \mathbf{t}) =  \sum_{i=0}^{d-1} [\mathbf{r}]_i \sum_{k=0}^{d-1} [\mathbf{h}]_k . [\mathbf{t}]_{(k+i)\mod d}
\end{equation}

Due to the use of circular correlation, HolE achieves $\mathcal{O} (n d + m d)$ space complexity and $\mathcal{O}(n_t d \log d)$ time complexity.

\subsection{Visualizing KG Embeddings}
We selected 10 driving segments (including their samples) from each dataset - NuScenes and Lyft - and created ``mini" KGs to visualize the embeddings in 2-dimensional (2D) space. After experimenting with both PCA \cite{wold1987principal} and t-Distributed Stochastic Neighbor Embedding (t-SNE)\cite{maaten2008visualizing}, t-SNE was selected for dimensionality reduction as its 2D projections yielded more meaningful clusters than PCA for the generated embeddings. The embedding dimension $d$ was set to 100 when generating embeddings for all our experiments.

Our extended Scene Ontology identifies \textit{events} and \textit{features-of-interests (FoI)} as top-level classes in the ontology, and each instance of \textit{event} or \textit{FoI} are linked to Scenes via the \textit{includes} relation. \textit{FoIs} are related to \textit{events} through the \textit{isParticipantOf} relation. Therefore, we first look at how FoIs and events are manifested in the embedding space for each dataset. 

\subsubsection{NuScenes KG Embeddings}
\begin{figure}[!htbp]
\centerline{\includegraphics[width=0.5\textwidth]{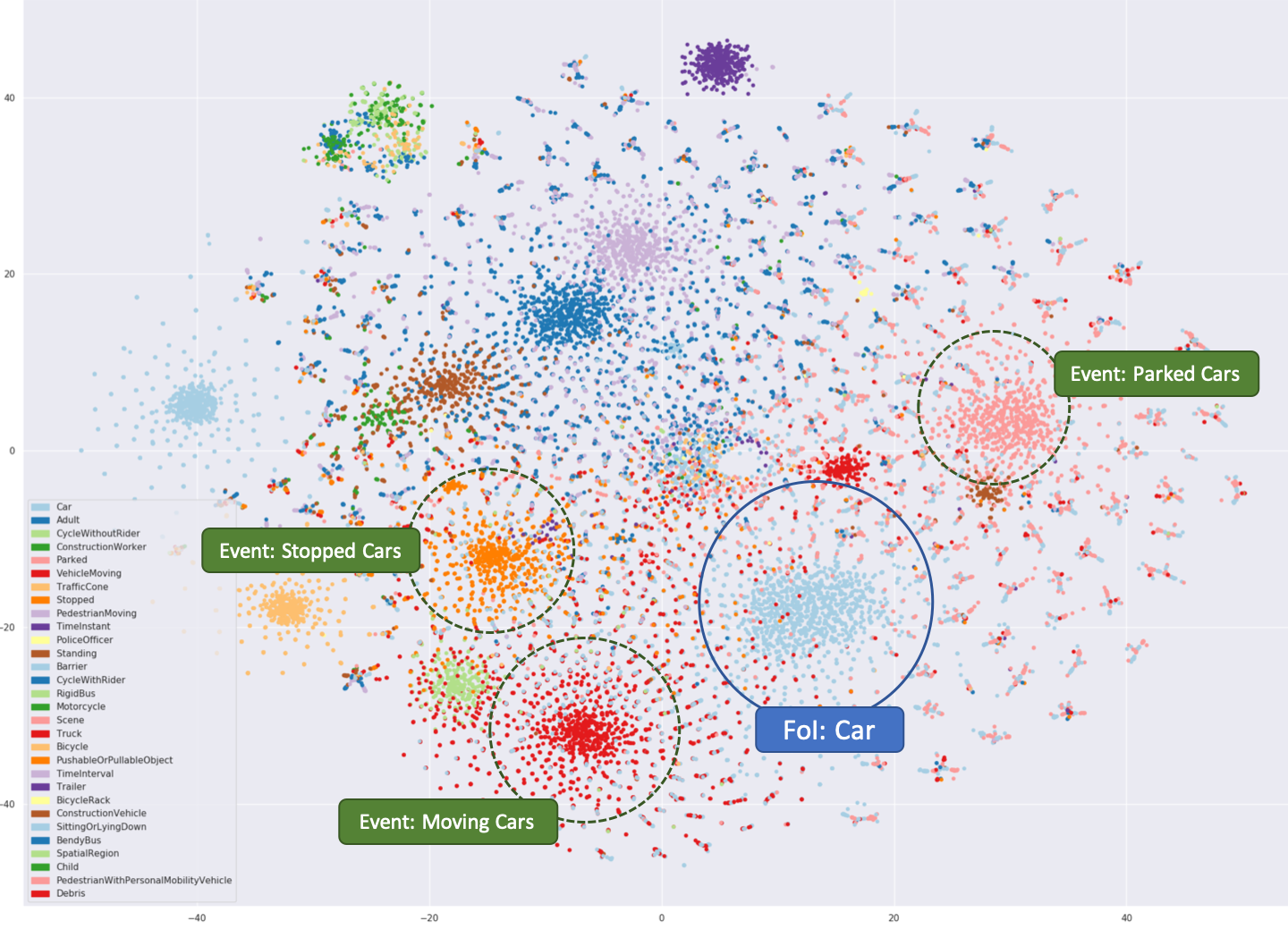}}
\caption{Clustering of FoIs with events in NuScenes dataset}
\label{fig:feature_events_nusc}
\end{figure}

Figure \ref{fig:feature_events_nusc} shows how the events and FoIs form clusters in the embedding space based on their type. For the sake of brevity, only \textit{cars} and the events in which they participate are highlighted. From this visualization, you can see that instances of events such as \textit {stopped car}, \textit {moving car} and \textit {parked car} are clustered around the instances of \textit{car}. The embeddings represented in this figure are generated from TransE on the ``Base KG".

\subsubsection{Lyft KG Embeddings}
\begin{figure}[h]
\centerline{\includegraphics[width=0.5\textwidth]{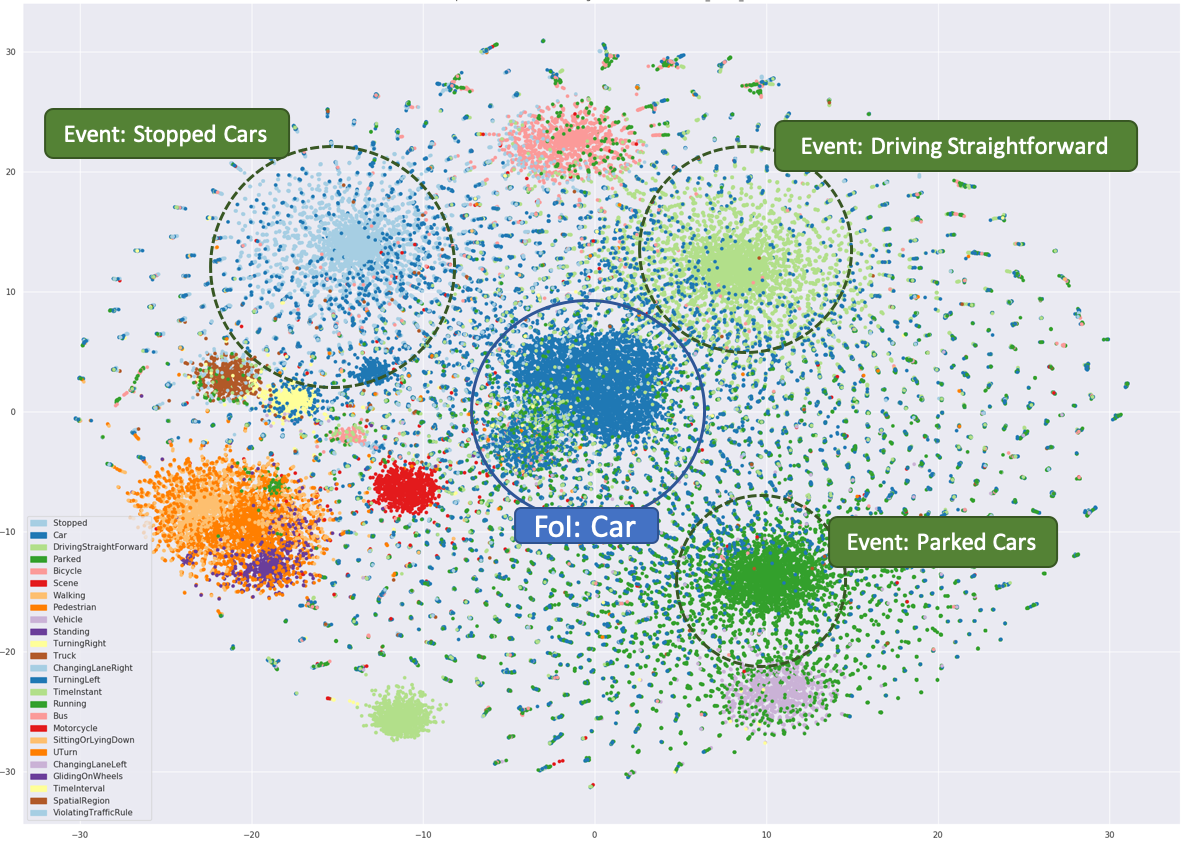}}
\caption{Clustering of FoIs with events in Lyft dataset}
\label{fig:feature_events_lyft}
\end{figure}

Figure \ref{fig:feature_events_lyft} shows a similar visualization of the embeddings from the Lyft ``Base KG". This image shows how the events in Lyft are clustered together with FoIs. It may be noticed that Lyft contains fewer clusters than NuScenes. This is the case since the Lyft dataset only contains annotations for a few FoIs.  Similar to what we've seen with NuScenes, Lyft embeddings also show how the instances of events such as \textit{stopped car}, \textit{parked car} and \textit{driving straightforward} are clustered around instances of \textit{car}.

\section{Evaluation}
The primary objective of our evaluation is to determine how well the salient features and rich semantics of KGs, such as type and relation semantics, transfer to learned embeddings. Our evaluation deviates from most prior work evaluating KGE algorithms, which focus on evaluating the performance of some extrinsic downstream task (such as entity classification). Here, we focus more on an intrinsic evaluation of embeddings. 

\subsection{Evaluation of KG Embeddings}
\begin{figure}[htbp]
\centerline{\includegraphics[width=0.45\textwidth]{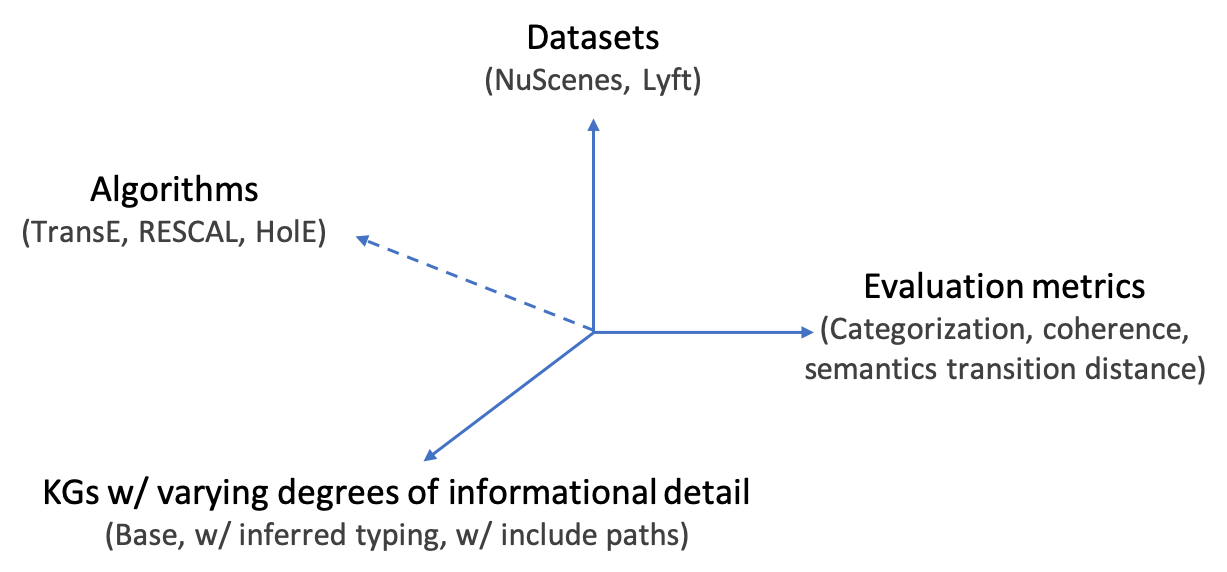}}
\caption{Four dimensions involved in our evaluation}
\label{fig:eval}
\end{figure}

There exists a large body of literature that evaluates the effect of using KGEs on a downstream task. To our knowledge, however, there's only one recent work which introduces metrics to evaluate and quantify the intrinsic quality of embeddings. Of the metrics introduced in \cite{alashargi2019metrics}, we adapt three metrics for our evaluation: \textit {categorization measure, coherence measure}, and \textit{semantic transition distance}. Figure \ref{fig:eval} depicts the four dimensions involved in our evaluation  -- i.e. quality metrics, datasets, KGE algorithms and KGs with varying degrees of informational detail.

Next we provide a brief overview of each of the evaluation metrics;

The \textbf{categorization measure} captures how well the entities that are ``typed" by the same background concept cluster together. For example, all the entities that are typed by the concept \textit{Car} ($\forall e_i \in c_{k=Car}$) share common characteristics and should be clustered together. Hence this metric is computed by taking the cosine similarity $s (V_1, V_2) = \frac{V_1. V_2}{|V_1||V_2|}$ of the averaged embedding vector (equation \ref{eq:avg}) of all such entities and the embedding vector of the background concept $k$, $\mathbf{c_k}$. 

\begin{equation}
    \forall e_i \in c_k,  \mathbf{\overline{e}_k} = \frac{1}{n} \sum_{i = 1}^{i = n} \mathbf{e_i}
    \label{eq:avg}
\end{equation}

\begin{equation}
    Categorization (\mathbf{c_k}) = s (\mathbf{\overline{e}_k, c_k})
    \label{eq:cat-measure}
\end{equation}

The \textbf{coherence measure} captures whether the adjacent entities in the embedding space share a common background concept. In introducing this measure, authors hypothesise that in the ideal case, all the entities that are typed by the same background concept should form a cluster and the background concept should be the centroid of this cluster. This is quantified (equation \ref{eq:coherence}) by taking $n$ closest entities of the background concept $c_i$ and taking the proportion of which that are actually typed by $c_i$. For our experiments, we choose $n$ to be 1000.

\begin{equation}
    Coherence (\mathbf{c_i}) = \frac{{\# \mathbf{e_i} |\mathbf{e_i} \in c_i }}{n} 
    \label{eq:coherence}
\end{equation}

The \textbf{semantic transition distance} is a widely used metric in word embedding literature and re-introduced to KGEs to capture the relational semantics of KGs. For example, assume $h_i$ is asserted as the domain of the property $r_i$ and $t_i$ is asserted as its range. Then, if ($h_i,r_i,t_i$) is correctly represented in the embedding space, the transition distance between $\mathbf{h_i} + \mathbf{r_i}$ should be close to $\mathbf{t_i}$. This is formally represented in equation \ref{eq:semantic-td} where $T_r$ is semantic transition distance of relation $r$ and $s$ denotes cosine similarity.

\begin{equation}
    T_r (h_i + r_i, t_i) = s (\mathbf {h_i + r_i, t_i})
    \label{eq:semantic-td}
\end{equation}

Next we report our evaluation results of these three metrics with respect to each dataset/algorithm.

\subsubsection{Evaluation on the Lyft Dataset}

\begin{figure}[htbp]
\centerline{\includegraphics[width=0.5\textwidth]{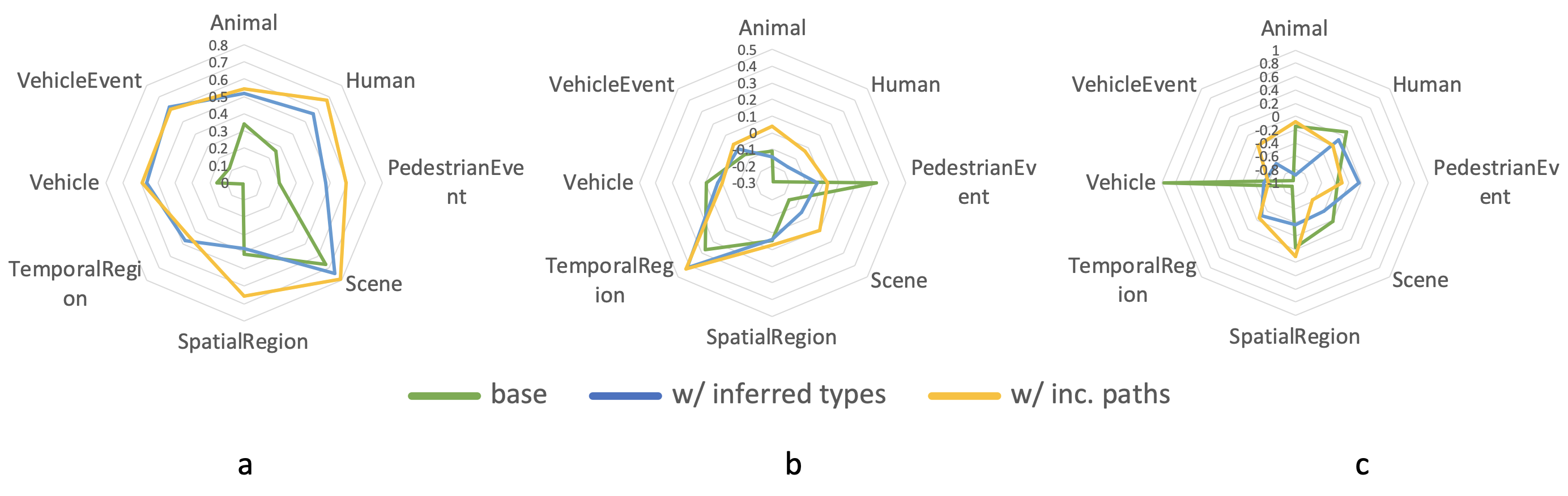}}
\caption{Categorization measure computed for different KGs on the embeddings generated from (a) TransE, (b) RESCAL and (c) HolE}
\label{fig:lyft_cat}
\end{figure}

Figure \ref{fig:lyft_cat} summarizes the results of the categorization measure computed on the embeddings generated from three algorithms on three KG versions. It is clear from the figures that TransE performs better compared to RESCAL and HolE and the categorization quality is mostly better in the KG with more informational detail (i.e. KG w/ include paths) compared to other two less expressive variants. In our experimental setting, this measure is computed considering the top level concepts (FoIs and events) in the Scene Ontology.

\begin{figure}[htbp]
\centerline{\includegraphics[width=0.5\textwidth]{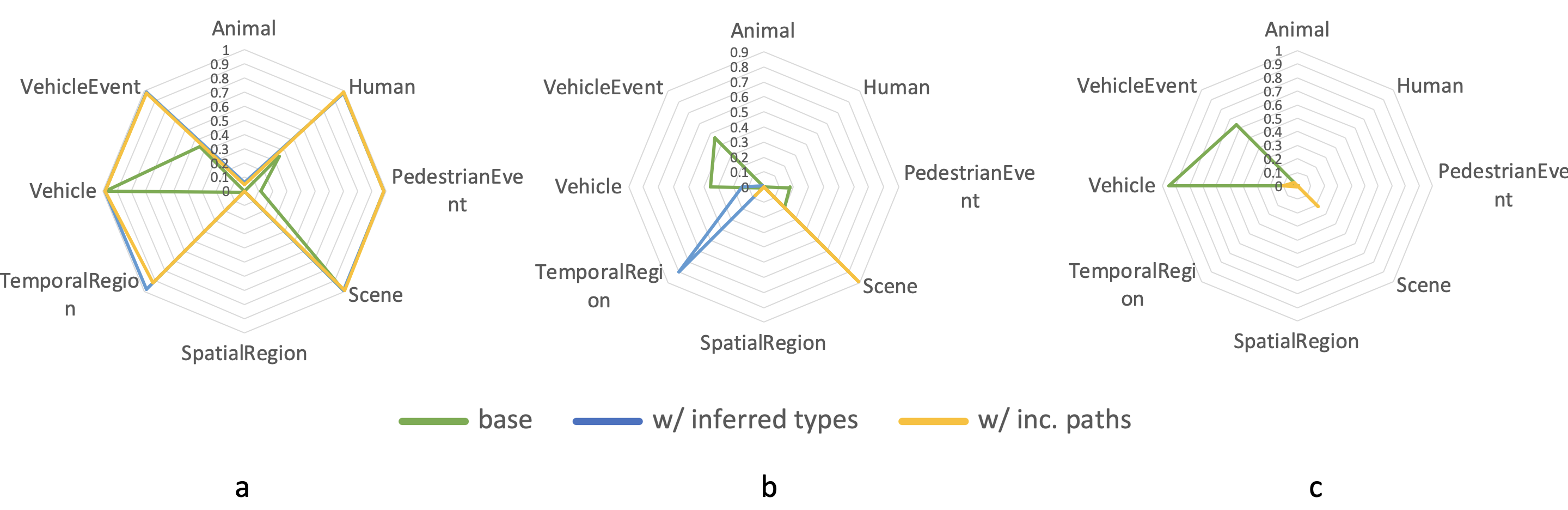}}
\caption{Coherence measure computed for different KGs on the embeddings generated from (a) TransE, (b) RESCAL and (c) HolE}
\label{fig:lyft_coh}
\end{figure}

The results of the coherence measure, as depicted in figure \ref{fig:lyft_coh},  shows a similar trend as the categorization measure; TransE is performing better and both RESCAL and HolE fail to generate entity clusters with high purity and closer to the background concept of those entities. It is interesting to note that, the KG with highest informational detail shows significant improvement in coherence measure on the embeddings generated from TransE.

\begin{figure}[!htbp]
\centerline{\includegraphics[width=0.5\textwidth]{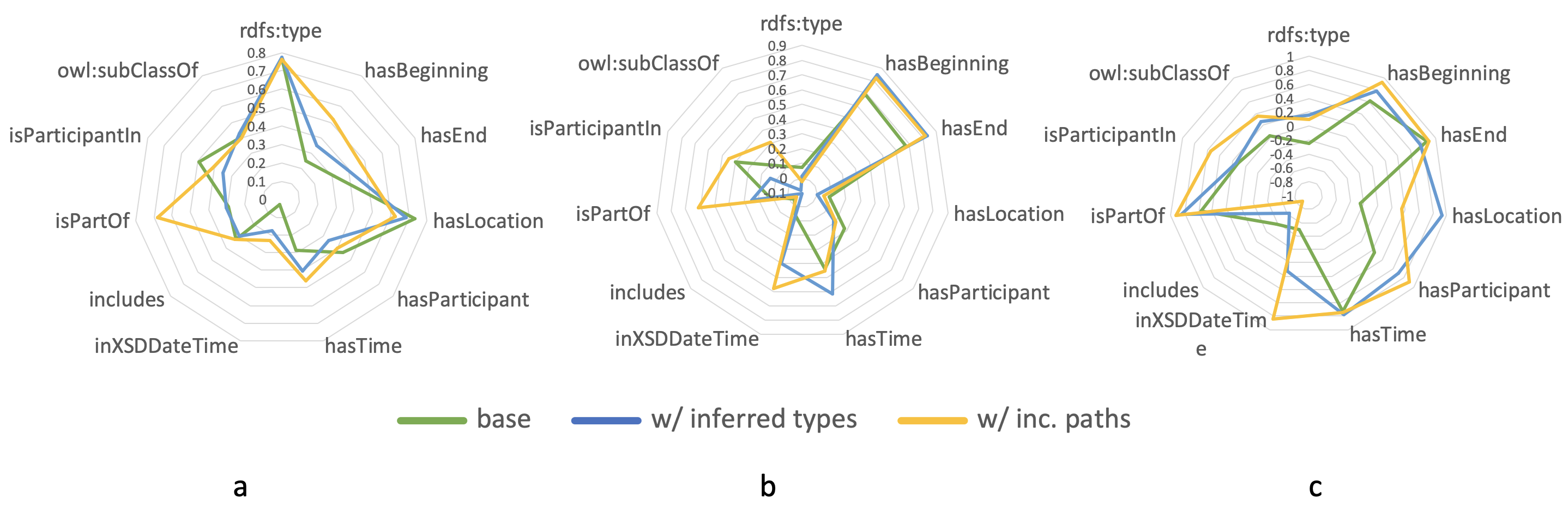}}
\caption{Semantic transition distance computed for different KGs on the embeddings generated from (a) TransE, (b) RESCAL and (c) HolE}
\label{fig:lyft_rel}
\end{figure}

Next we look at how the relational semantics in KGs are transferred to KGEs by computing semantic transition distance for 11 relations defined in the Scene ontology. As per figure \ref{fig:lyft_rel}, KGs with include paths are able to capture relational semantics better than the other two variants across all three algorithms. An interesting observation to note here is that the \textit{isPartOf} relation performs significantly better in KGs with include paths across all three algorithms even though we have only added implicit \textit{include} relations. A possible explanation could be that the implicit \textit{include} paths make the relationship between scenes and sub-scenes stronger in KGs with the highest informational detail.

\subsubsection{Evaluation on the NuScenes Dataset}
The evaluation process on the NuScenes dataset is similar to Lyft. However, we report that RESCAL was not scalable to the NuScenes KGs (having 10.8+ million triples and 2.1+ million entities). Therefore, we evaluate NuScenes KGEs only on TransE and HolE. 

\begin{figure}[!ht]
\centerline{\includegraphics[width=0.5\textwidth]{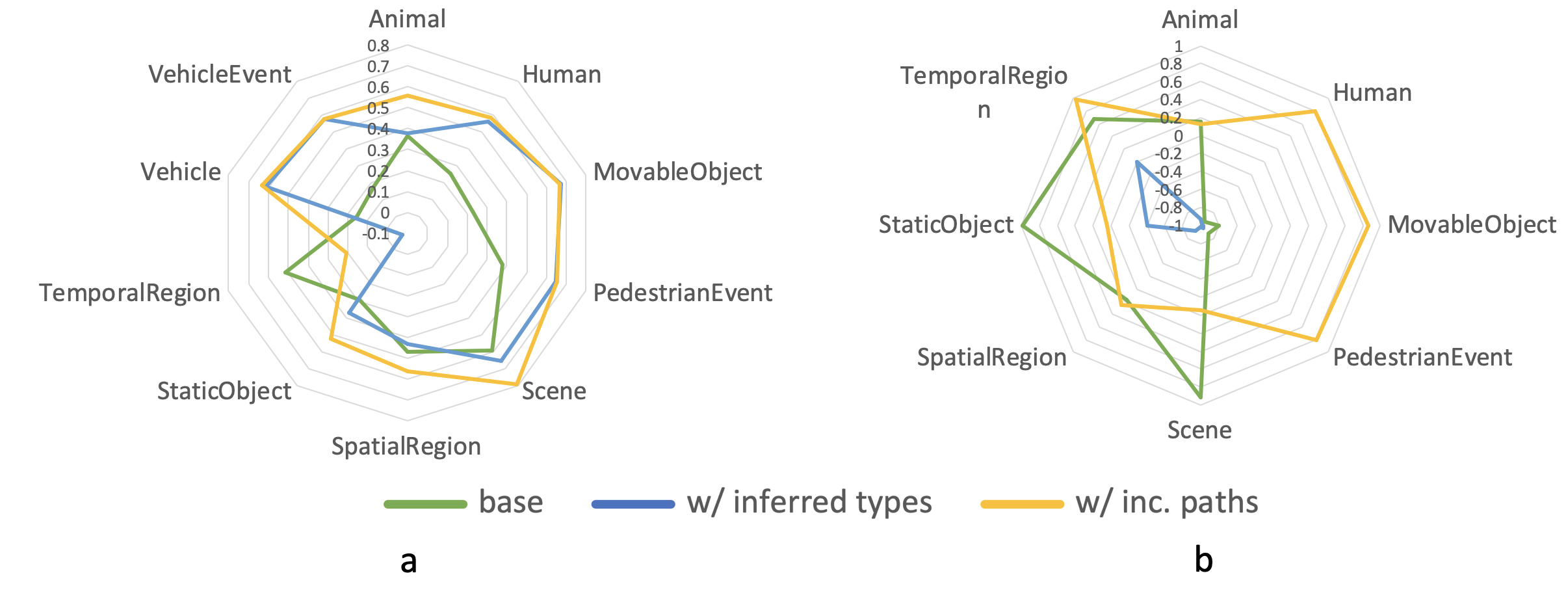}}
\caption{Categorization measure computed for different KGs on the embeddings generated from (a) TransE and (b) HolE}
\label{fig:nusc_cat}
\end{figure}

The results of the categorization measure for the NuScenes dataset follows a similar trend as Lyft (see Figure \ref{fig:nusc_cat}). TransE embeddings on KG w/ include paths yields the best categorization performance, with the  exception of a few concepts where HolE outperforms on the base KG. Except for these few outliers, the results show that higher level of informational detail in KG achieves better categorization irrespective of the KGE algorithms used for training.

\begin{figure}[!ht]
\centerline{\includegraphics[width=0.5\textwidth]{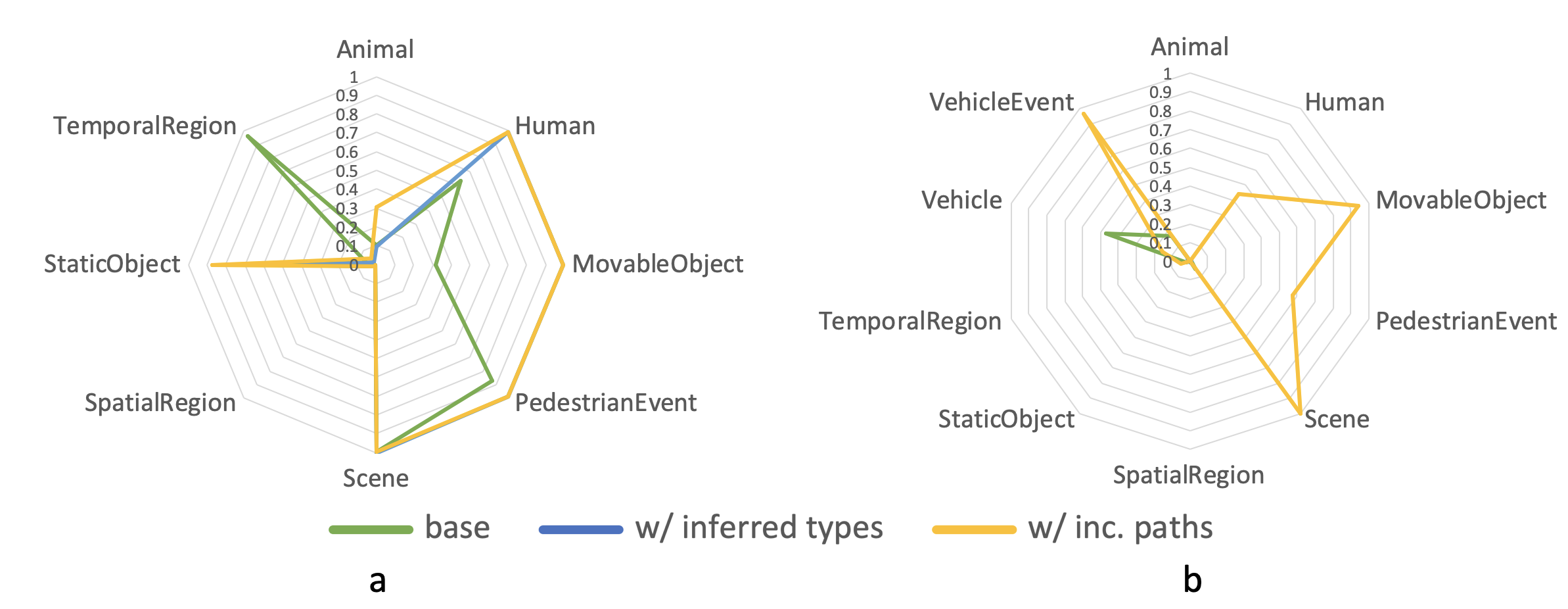}}
\caption{Coherence measure computed for different KGs on the embeddings generated from (a) TransE and (b) HolE}
\label{fig:nusc_coh}
\end{figure}

The benefits of information detail in KG are portrayed well in the results of coherence measure (see Figure \ref{fig:nusc_coh}). Even though the coherence measure, for many concepts, are either non-existent or closer to zero, the KG w/ include paths significantly outperform the other two KG variants for those values that exist.

\begin{figure}[!ht]
\centerline{\includegraphics[width=0.5\textwidth]{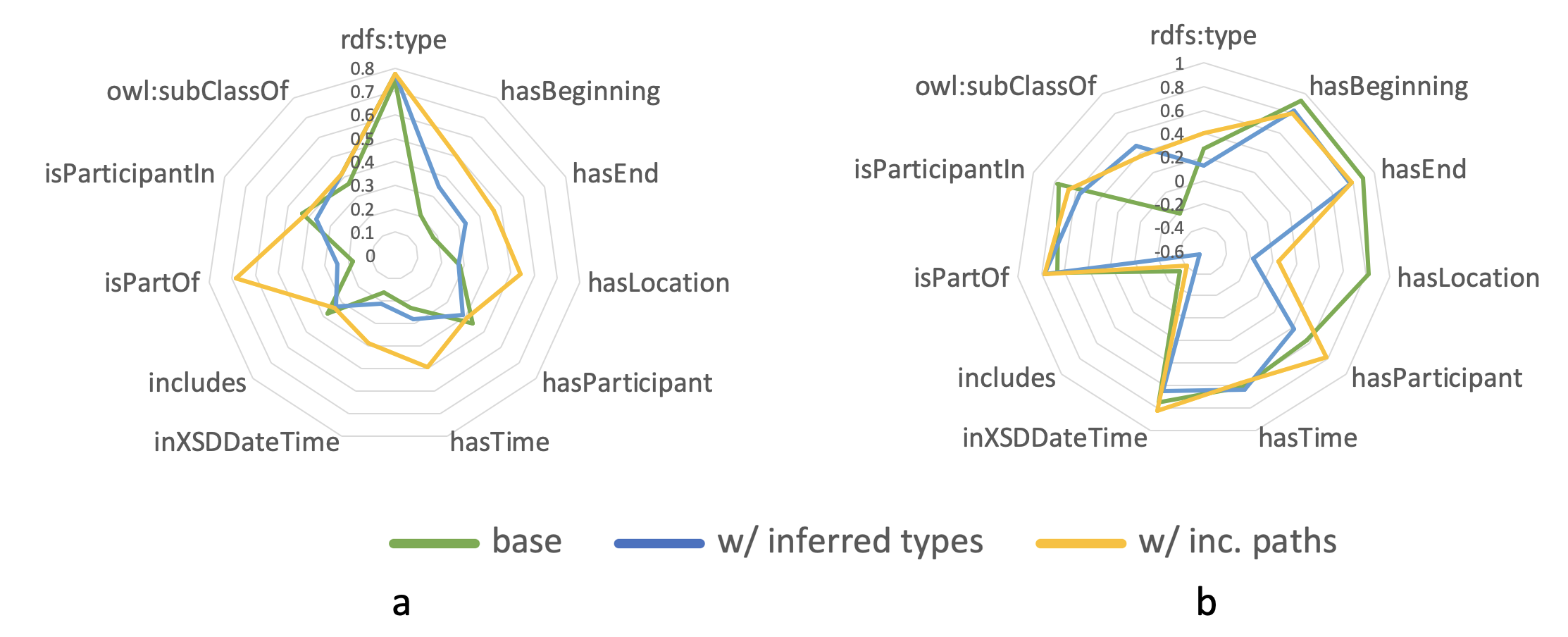}}
\caption{Semantic transition distance computed for different KGs on the embeddings generated from (a) TransE and (b) HolE}
\label{fig:nusc_rel}
\end{figure}

The results of the semantic transition distance for TransE show consistent patterns similar to Lyft (see Figure \ref{fig:nusc_rel}). HolE, however, shows that base KGEs perform on par with embeddings trained on the KG w/ include paths.

\subsection{Discussion}
The evaluation of KGEs for AD domain lead to some interesting observations. We discuss our observations in three perspectives: (1) KGE algorithmic perspective, (2) evaluation measures, and (3) various levels of KG informational detail. 

First, looking at the overall performance of KGE algorithms, TransE performs better than RESCAL and HolE in capturing both type and relational semantics. TransE is also scalable to large KGs and shows consistent performance across datasets. In addition to RESCAL's space and time complexity, it's performance on all three metrics is worse than TransE and HolE. Even though HolE's performance is sub-optimal compared to TransE, it was consistent across the two datasets and the derived KGs. We hypothesize that the better performance of TransE on all three metrics is due to the way it is learning embeddings; i.e. using the translational distance based scoring function inspired by word embedding algorithms. The KGE quality metrics introduced by \cite{alashargi2019metrics} are inspired by the word embedding literature. They evaluated these metrics on KGEs generated from word embedding based RDF2Vec \cite{ristoski2016rdf2vec} algorithm. Hence, it may be worth examining whether these metrics are suitable only for evaluating embeddings generated from translational distance / word embedding inspired KGE algorithms.

Second, when considering the evaluation measures used, we observe that the coherence measure is not very meaningful in this domain to evaluate the quality of KGEs. The entities are mostly clustered based on either scenes/sub-scenes or FoIs/events. Hence, the $n$ most similar entities to a class (e.g. \textit{Human}) are mostly not homogeneous, resulting in zero or close to zero coherence value. 

Third, it has been consistently shown across multiple datasets and algorithms that KGs with the highest levels of informational detail are able to capture both type and relational semantics better than the other two less expressive variants. This discovery leads to an interesting future direction for research. To the best of our knowledge, all existing KGE algorithms in the literature are evaluated on base KGs (i.e. KGs without any inference). Therefore, it stands to reason that a KG embedding derived from a KG with more informational detail should capture more salient features and rich semantics of the KG. Such informational details can be captured either through pre-processing or by automatically extracted by the KGE algorithm.

\section {Investigating Semantics of AD Domain}
In the previous section we looked at different quality aspects of embeddings which are common to KGs in any domain. Here, we report additional experiments which look at certain aspects specific to AD domain and scene understanding.

\subsection{Scene/sub-scene Relationship}
An understanding of complex AD scenes is an important task in AD domain. The ability to distinguish one complex scene from another requires looking at (1) how the scene/sub-scene relationship (formally defined by \textit{isPartOf} relation) is captured by the embeddings in different KG versions, and (2) how well the FoIs and events cluster based on the scenes/subscenes they belong to. 

\begin{figure}[!htbp]
\centerline{\includegraphics[width=0.5\textwidth]{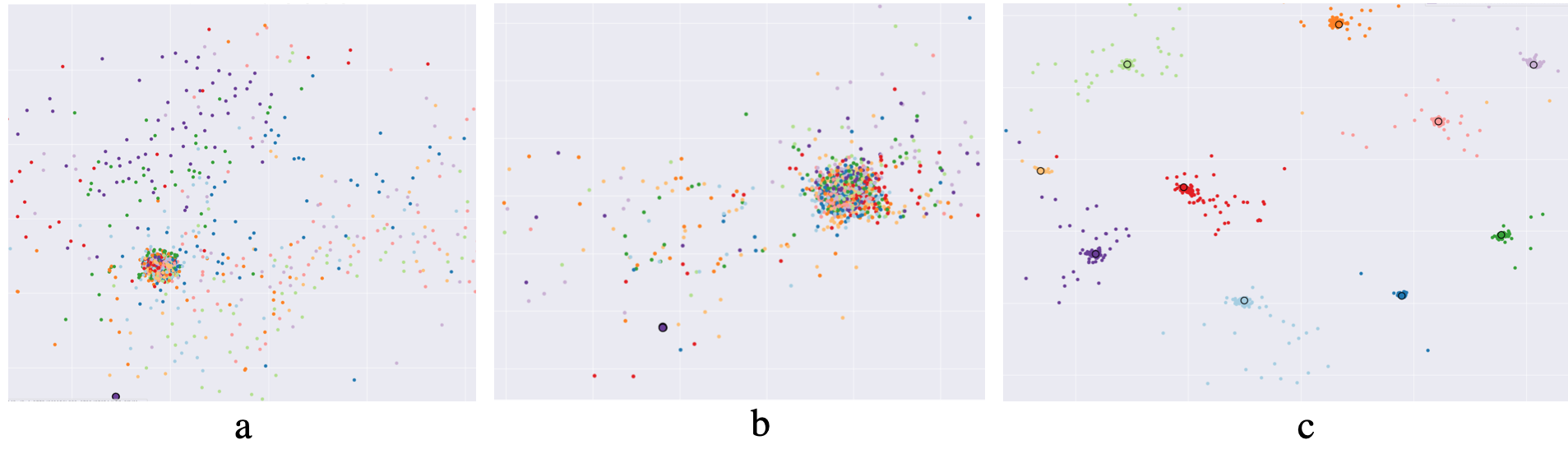}}
\caption{Clustering of scenes with sub-scenes in different KG versions of Lyft: (a) base KG, (b) KG w/ inferred types, and (c) KG w/ include paths}
\label{fig:scene-subscene}
\end{figure}

Figure \ref{fig:scene-subscene} shows the manifestation of the scene-subscene relationship that is moving from a ``no-relation" (figure \ref{fig:scene-subscene}(a)) in the base KG to a more ``meaningful" one (figure \ref{fig:scene-subscene}(c)) in the KG with the highest informational detail. Then we look at how the various levels of informational detail in KGs affect the clustering of FoIs and events of a scene based on scene/sub-scene relationship. 

\begin{figure}[!htbp]
\centerline{\includegraphics[width=0.5\textwidth]{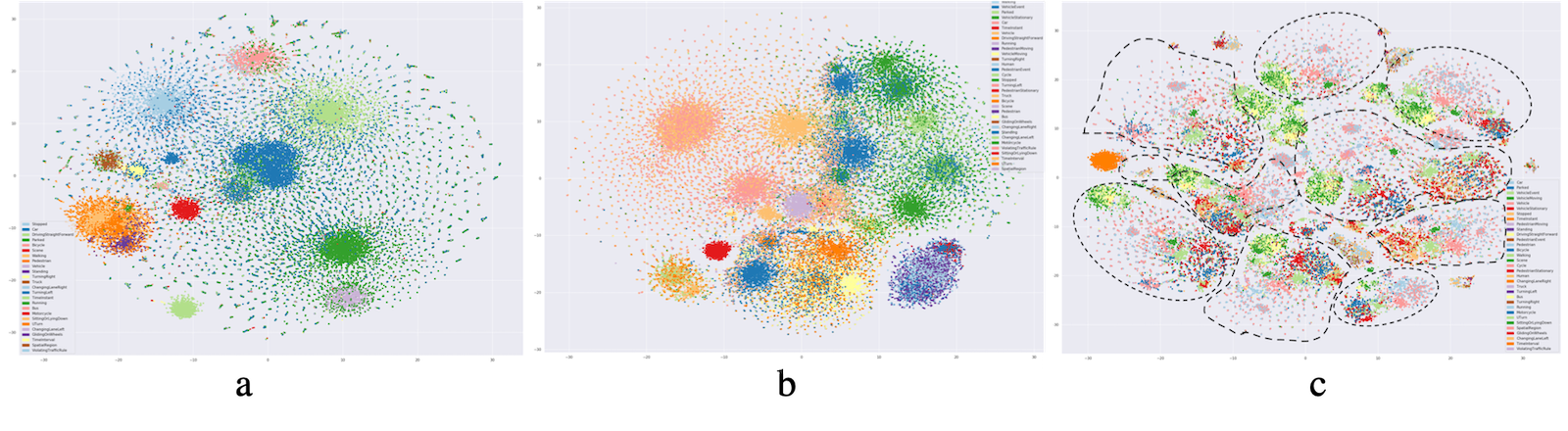}}
\caption{Clustering of FoIs and events together with scenes/sub-scenes in different KG versions: (a) base KG, (b) KG w/ inferred types, and (c) KG w/ include paths}
\label{fig:features-events-scenes}
\end{figure}

Figure \ref{fig:features-events-scenes} shows how the FoI/event dominant clusters in the base KG transfer to a clustering based on 10 scenes in the KG w/ include paths. Interestingly, we can still see small clusters formed based on FoIs/events inside the larger scene clusters. This suggests that the KGs with access to more informational detail are able to distinguish a scene by both participating FoIs/events as well as scene/sub-scene relationships. 

\subsection{KGEs for Computing Scene Similarity}
\begin{figure}[!htbp]
\centerline{\includegraphics[width=0.5\textwidth]{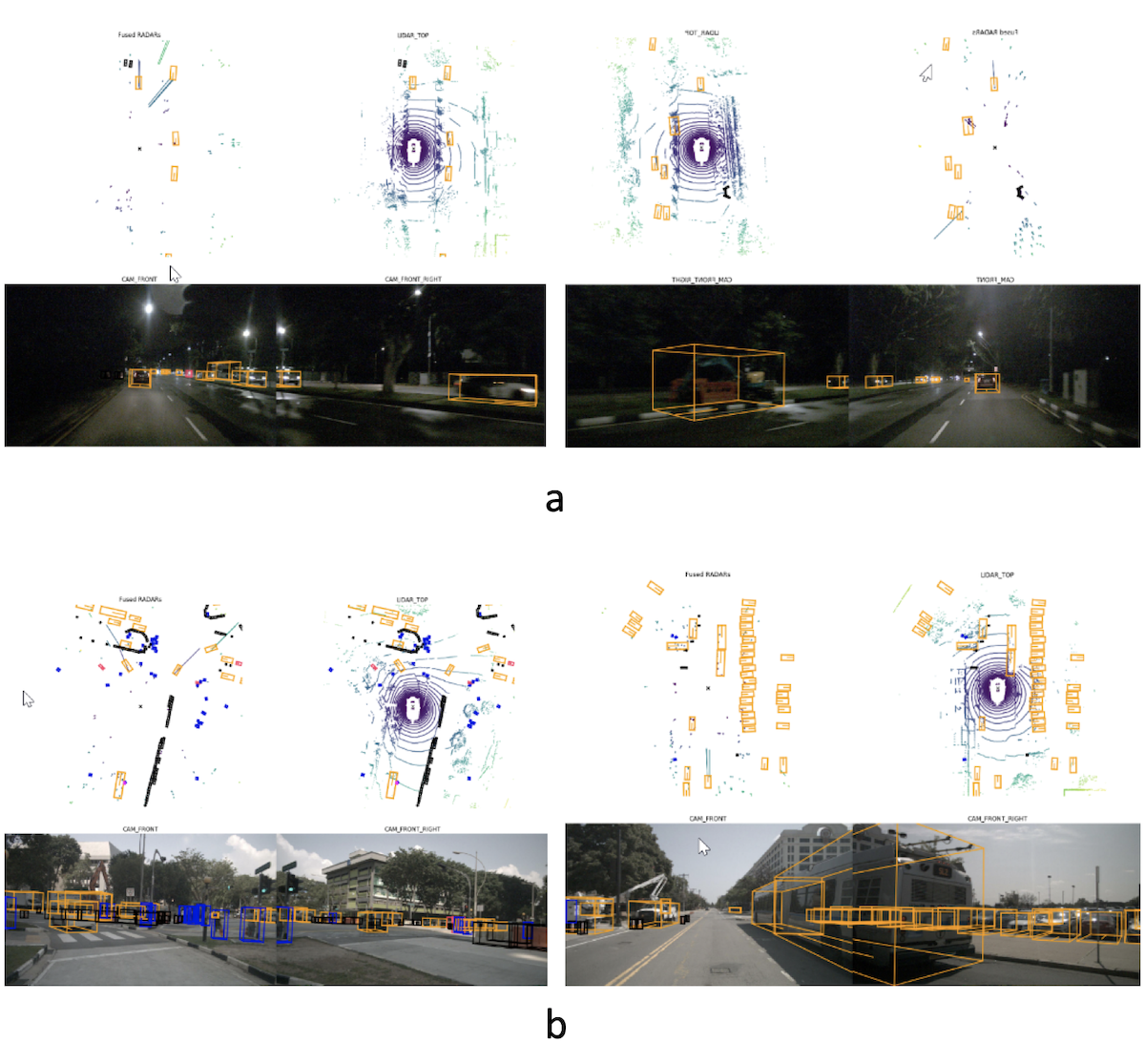}}
\caption{Most similar sub-scenes computed using KGEs trained on NuScenes Base KG; (a) sub-scenes from the same scene and (b) sub-scenes from two different scenes}
\label{fig:scene-similarity}
\end{figure}

We report preliminary results of using KGEs for computing scene similarity. Our objective here is to determine whether two scenes are similar by considering only KGEs. Given a set of scene pairs, we calculate the cosine similarity between the KGE vectors of two scenes in a pair and then select pairs with the highest cosine similarity. Figure \ref{fig:scene-similarity}(a) shows the two most similar sub-scenes when pairs include sub-scenes from the same parent scene. With further investigation, we found that these two sub-scenes are in fact subsequent frames (or samples) from the same 20 second driving segment. Figure \ref{fig:scene-similarity}(b) shows the two most similar sub-scenes when pairs contain only sub-scenes from different scenes. It is interesting to note that the KGE based similarity computation was able to identify two sub-scenes which are not visually similar, but share common characteristics. For example, the black string of objects in Figure \ref{fig:scene-similarity}(b) (Left) are barriers (a \textit{Static Object}) and the orange string of objects in Figure \ref{fig:scene-similarity}(b) (Right) are set of stopped cars.

\section{Conclusion}
In this paper, we present an evaluation of KGEs for the autonomous driving domain that considers multiple datasets, metrics, algorithms and levels of informational detail. This evaluation supports the hypothesis that a KG with more detailed information yields higher quality KG embeddings with respect to both type and relational semantics. Furthermore, this evaluation highlights an important question about the suitability of metrics used in the existing literature, to evaluate a wide range of KGE algorithms. Finally, opening rich areas for future research, we present an early investigation into the use of KGEs for two important use-cases from the AD domain: scene/sub-scene understanding and computing scene similarity. 

\bibliography{aaai-make.bib}
\bibliographystyle{aaai}

\section*{Appendices}
We include detailed evaluation results with respect to each dataset, evaluation metric, KG and algorithm in appendices. Tables (\ref{tab:lyft-cat}, \ref{tab:lyft-coh}, \ref{tab:lyft-rel}) in appendix A contains results of the Lyft dataset whereas appendix B contains tables (\ref{tab:nusc-cat}, \ref{tab:nusc-coh}, \ref{tab:nusc-rel}) summarizing the results of the NuScenes dataset.

\onecolumn 
\section*{Appendix A: Evaluation Results of the Lyft dataset}
\begin{table}[!h]
\centering
\begin{tabular}{l|lll|lll|lll}

\multicolumn{1}{c}{} & \multicolumn{3}{c}{TransE } &  \multicolumn{3}{c}{RESCAL } & \multicolumn{3}{c}{HolE} \\ \hline
Class & base & w/ types &  w/ paths & base & w/ types &  w/ paths & base & w/ types &  w/ paths \\\hline
Animal & 0.3404 & 0.5192 & 0.5420 & -0.1072 & -0.1420 & 0.0405 & -0.1465 & -0.8855 & -0.0796\\
Human & 0.2570 & 0.5644 & 0.6767 & -0.2916 & -0.1684 & -0.0281 & 0.0867 & -0.0756 & -0.2007\\
PedestrianEvent & 0.2057 & 0.4691 & 0.5891 & 0.3228 & -0.0310 & 0.0271 & -0.3618 & -0.0415 & -0.2968\\
Scene & 0.6654 & 0.7390 & 0.7904 & -0.1591 & -0.0511 & 0.1011 & -0.1861 & -0.3845 & -0.6410\\
SpatialRegion & 0.4127 & 0.3791 & 0.6552 & 0.0443 & 0.0373 & 0.0735 & -0.0349 & -0.3709 & 0.1069\\
TemporalRegion & 0.0075 & 0.4787 & 0.4500 & 0.2709 & 0.4206 & 0.4332 & -0.9270 & -0.2917 & -0.2464\\
Vehicle & 0.1604 & 0.5616 & 0.5898 & 0.0997 & 0.0286 & 0.0045 & 0.9681 & -0.5235 & -0.5934\\
VehicleEvent & 0.1173 & 0.6167 & 0.6063 & -0.0617 & -0.0114 & 0.0259 & -0.9617 & -0.5908 & -0.2173\\
\hline
\end{tabular}
\caption{Results of categorization measure computed on embeddings generated from TransE, RESCAL and HolE}
\label{tab:lyft-cat}
\end{table}

\begin{table}[!h]
\centering
\begin{tabular}{l|lll|lll|lll}

\multicolumn{1}{c}{} & \multicolumn{3}{c}{TransE} &  \multicolumn{3}{c}{RESCAL } & \multicolumn{3}{c}{HolE} \\ \hline
Class & base & w/ types &  w/ paths & base & w/ types &  w/ paths & base & w/ types &  w/ paths \\\hline
Animal & 0.002 & 0.062 & 0.047 & 0 & 0 & 0 & 0 & 0 & 0\\
Human & 0.346 & 0.986 & 0.991 & 0.001 & 0 & 0 & 0 & 0.002 & 0\\
PedestrianEvent & 0.117 & 0.983 & 0.986 & 0.176 & 0 & 0 & 0.001 & 0 & 0\\
Scene & 0.996 & 0.996 & 0.998 & 0.193 & 0.674 & 0.89 & 0.004 & 0.003 & 0.215\\
SpatialRegion & 0.001 & 0.001 & 0.001 & 0 & 0 & 0 & 0 & 0 & 0\\
TemporalRegion & 0.002 & 0.982 & 0.917 & 0.001 & 0.794 & 0.077 & 0 & 0 & 0.001\\
Vehicle & 0.987 & 0.99 & 0.99 & 0.352 & 0.146 & 0 & 0.947 & 0 & 0.108\\
VehicleEvent & 0.445 & 0.987 & 0.986 & 0.46 & 0.009 & 0 & 0.642 & 0 & 0.036\\
\hline
\end{tabular}
\caption{Results of coherence measure computed on embeddings generated from TransE, RESCAL and HolE}
\label{tab:lyft-coh}
\end{table}

\begin{table}[!h]
\centering
\begin{tabular}{l|lll|lll|lll}

\multicolumn{1}{c}{} & \multicolumn{3}{c}{TransE } &  \multicolumn{3}{c}{RESCAL } & \multicolumn{3}{c}{HolE} \\ \hline
Relation & base & w/ types &  w/ paths & base & w/ types &  w/ paths & base & w/ types &  w/ paths \\\hline
rdfs:type & 0.7650 & 0.7744 & 0.7626 & 0.0722 & 0.0119 & -0.0243 & -0.2436 & 0.1521 & 0.0890\\
hasBeginning & 0.2483 & 0.3510 & 0.5145 & 0.6912 & 0.8491 & 0.8285 & 0.6179 & 0.7860 & 0.9338\\
hasEnd & 0.3134 & 0.3939 & 0.4732 & 0.6720 & 0.8314 & 0.8203 & 0.8675 & 0.7588 & 0.9001\\
hasLocation & 0.7310 & 0.6824 & 0.6261 & 0.0877 & 0.0048 & 0.0481 & -0.2506 & 0.9296 & 0.3473\\
hasParticipant & 0.4392 & 0.3416 & 0.4010 & 0.2772 & 0.1781 & 0.2010 & 0.2363 & 0.6978 & 0.8979\\
hasTime & 0.2883 & 0.4057 & 0.4601 & 0.4375 & 0.6180 & 0.4592 & 0.7264 & 0.7720 & 0.7412\\
inXSDDateTime & 0.0260 & 0.1786 & 0.2335 & 0.0589 & 0.4005 & 0.5822 & -0.4892 & 0.1231 & 0.8463\\
includes & 0.3271 & 0.3051 & 0.3327 & -0.0341 & -0.0945 & -0.0537 & -0.3785 & -0.6282 & -0.8829\\
isPartOf & 0.2893 & 0.3039 & 0.6799 & 0.1291 & 0.2470 & 0.6166 & 0.5646 & 0.8461 & 0.9255\\
isParticipantIn & 0.4964 & 0.3480 & 0.4121 & 0.3967 & 0.1403 & 0.4515 & 0.1110 & 0.1449 & 0.5452\\
owl:subClassOf & 0.4036 & 0.4253 & 0.3969 & 0.1133 & -0.0816 & 0.3055 & 0.0297 & 0.2703 & 0.3664\\
\hline
\end{tabular}
\caption{Semantic transition distance computed on embeddings generated from TransE, RESCAL and HolE}
\label{tab:lyft-rel}
\end{table}

\onecolumn 
\section*{Appendix B: Evaluation Results of the NuScenes dataset}

\begin{table}[!h]
\centering
\begin{tabular}{l|lll|lll}

\multicolumn{1}{c}{} & \multicolumn{3}{c}{TransE } &  \multicolumn{3}{c}{HolE } \\ \hline
Class & base & w/ types &  w/ paths & base & w/ types &  w/ paths \\\hline
Animal & 0.3649 & 0.3769 & 0.5565 & 0.1582 & -0.9250 & 0.1250\\
Human & 0.2530 & 0.5573 & 0.5798 & -0.9436 & -0.9865 & 0.7919\\
MovableObject & 0.2276 & 0.6694 & 0.6653 & -0.7962 & -0.9832 & 0.8592\\
PedestrianEvent & 0.3798 & 0.6447 & 0.6514 & -0.8769 & -0.9729 & 0.8101\\
Scene & 0.5931 & 0.6585 & 0.7950 & 0.9181 & -0.9856 & -0.0491\\
SpatialRegion & 0.4701 & 0.4313 & 0.5618 & 0.1635 & -0.9134 & 0.2596\\
StaticObject & 0.2914 & 0.3748 & 0.5274 & 0.9906 & -0.4043 & 0.0409\\
TemporalRegion & 0.5120 & -0.0722 & 0.2073 & 0.6798 & -0.0011 & 0.9696\\
Vehicle & 0.1556 & 0.6140 & 0.6352 & 0.7423 & -0.7442 & 0.8521\\
VehicleEvent & 0.1761 & 0.5740 & 0.5768 & -0.9159 & -0.5617 & -0.8291\\
\hline
\end{tabular}
\caption{Results of categorization measure computed on embeddings generated from TransE and HolE}
\label{tab:nusc-cat}
\end{table}

\begin{table}[!h]
\centering
\begin{tabular}{l|lll|lll}

\multicolumn{1}{c}{} & \multicolumn{3}{c}{TransE } &  \multicolumn{3}{c}{HolE } \\ \hline
Class & base & w/ types &  w/ paths & base & w/ types &  w/ paths \\\hline
Animal & 0.099 & 0.096 & 0.308 & 0 & 0 & 0\\
Human & 0.63 & 0.992 & 0.995 & 0.014 & 0.001 & 0.441\\
MovableObject & 0.315 & 0.994 & 0.995 & 0 & 0 & 0.944\\
PedestrianEvent & 0.875 & 0.995 & 0.994 & 0.005 & 0.003 & 0.574\\
Scene & 0.997 & 0.999 & 0.997 & 0.049 & 0 & 0.999\\
SpatialRegion & 0.004 & 0.004 & 0.004 & 0 & 0 & 0\\
StaticObject & 0.044 & 0.535 & 0.869 & 0 & 0 & 0\\
TemporalRegion & 0.964 & 0.019 & 0.041 & 0.022 & 0.001 & 0.048\\
Vehicle & 0.477 & 0.984 & 0.983 & 0.471 & 0.007 & 0.156\\
VehicleEvent & 0.917 & 0.993 & 0.992 & 0.165 & 0.021 & 0.965\\
\hline
\end{tabular}
\caption{Results of coherence measure computed on embeddings generated from TransE and HolE}
\label{tab:nusc-coh}
\end{table}

\begin{table}[!h]
\centering
\begin{tabular}{l|lll|lll}

\multicolumn{1}{c}{} & \multicolumn{3}{c}{TransE } &  \multicolumn{3}{c}{HolE } \\ \hline
Relation & base & w/ types &  w/ paths & base & w/ types &  w/ paths \\\hline
rdfs:type & 0.7426 & 0.7718 & 0.7733 & 0.2700 & 0.1280 & 0.3995\\
hasBeginning & 0.2040 & 0.3510 & 0.4977 & 0.9243 & 0.8186 & 0.7896\\
hasEnd & 0.1811 & 0.3344 & 0.4635 & 0.8960 & 0.7712 & 0.7919\\
hasLocation & 0.2823 & 0.2767 & 0.5411 & 0.8240 & -0.1710 & 0.0468\\
hasParticipant & 0.4417 & 0.3869 & 0.4057 & 0.5669 & 0.4233 & 0.7795\\
hasTime & 0.2325 & 0.2837 & 0.4973 & 0.6117 & 0.6332 & 0.5675\\
inXSDDateTime & 0.1650 & 0.2134 & 0.3917 & 0.7465 & 0.6430 & 0.8233\\
includes & 0.3767 & 0.3311 & 0.3404 & -0.3296 & -0.5496 & -0.4141\\
isPartOf & 0.1826 & 0.2498 & 0.6831 & 0.6641 & 0.7712 & 0.7725\\
isParticipantIn & 0.4307 & 0.3674 & 0.4203 & 0.7694 & 0.5585 & 0.6632\\
owl:subClassOf & 0.3646 & 0.4141 & 0.4131 & -0.2153 & 0.4608 & 0.3716\\

\hline
\end{tabular}
\caption{Semantic transition distance computed on embeddings generated from TransE and HolE}
\label{tab:nusc-rel}
\end{table}

\end{document}